# A New Benchmark Dataset for Texture Image Analysis and Surface Defect Detection

*Shervan Fekri-Ershad*
Faculty of Computer Engineering, Najafabad Branch, Islamic Azad University, Najafabad, Iran
Email: fekriershad@pco.iaun.ac.ir

***Abstract***— Texture analysis plays an important role in many image processing applications to describe the image content or objects. On the other hand, visual surface defect detection is a highly research field in the computer vision. Surface defect refers to abnormalities in the texture of the surface. So, in this paper a dual purpose benchmark dataset is proposed for texture image analysis and surface defect detection titled stone texture image (STI dataset). The proposed benchmark dataset consist of 4 different class of stone texture images. The proposed benchmark dataset have some unique properties to make it very near to real applications. Local rotation, different zoom rates, unbalanced classes, variation of textures in size are some properties of the proposed dataset. In the result part, some descriptors are applied on this dataset to evaluate the proposed STI dataset in comparison with other state-of-the-art datasets.

## I. Introduction

Texture, shape and color are main features that are used for to describe images. Texture information play an important role in many image processing applications such as medical image analysis [1], texture classification [2], object recognition[3] , motion tracking[4], noise reduction [5], inspection systems [6], etc.

Since now many descriptors have been proposed for texture analysis. In all cases, a benchmark dataset should be used to evaluate the performance of texture descriptors. Outex [7], Brodatz[8], Vistex[9], Trunk12 [10] and so on, are some of the well known texture image datasets. In near all of the texture datasets, textural images is gathering form different classes in terms of object, background, content, etc. Textural image referred to the image where a unique pattern is repeated whole the image with some little differences. Textural images can be categorized in two groups natural and intelligent. Natural textures are real world senses, where intelligent textures are images designed by human like wall or buildings.

Each abnormality on the surface of a product is known as defect. Any hole, crack, streaks, bump and so on are popular abnormalities. So, defect detection is a main aspect of quality control inspection in many factories. Sine now, many approaches have been proposed as visual inspection systems. In near all of them, surface defect detection is a basic aim. Usually, abnormalities on the product surfaces may disturb the repetitive texture of the surface. So, in many cases, texture image descriptors are used for surface defect detection in feature extraction phase. In this respect, benchmark datasets for defect detection should be included textural images.

In this respect, a dual purpose dataset is proposed in this paper for texture image analysis and surface defect detection.

The reminder of this paper is as follows: In section 2, the well known benchmark texture image analysis datasets are reviewed with details. In section 3, our collected dataset is proposed. In section 4, comparisons results of apply texture descriptors on proposed STI dataset is proposed. Finally conclusion is included.

## II. Related Benchmark Datasets

Each dataset have some unique properties which make it useful for special applications. In this section benchmark datasets for texture analysis is reviewed with details.

### *A. Brodatz*

Brodatz [8] is one of the most popular and all-purpose datasets that is included natural textures provides by Brodatz through photograph scanned after print. The Brodatz dataset has 112 texture images with resolution of 640×640 pixel and in 8 bit (256 gray values). The Brodatz album has some limitation like single illumination and viewing direction for each one of the texture. Some sample images of the Brodatz dataset are shown in Fig 1.

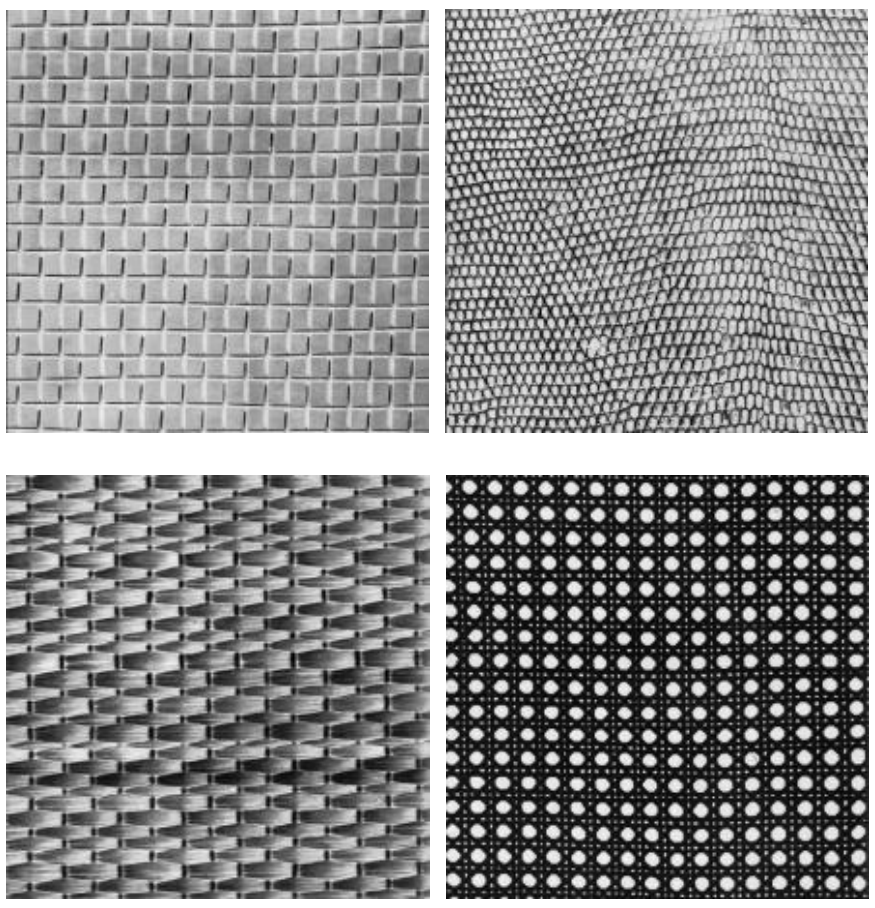

Figure 1. Some Examples of the Brodatz Dataset

*B. Outex*

Outex dataset was proposed by Oulue university researcher in 2002 [7]. Outex dataset is a most popular and largest dataset for texture analysis and texture classifications. This dataset has both the natural and artificial texture images, which with high applicability and open source availability. There are 16 different image sets in Outex. Among them, Tc_000010, Tc_000012 and Tc_000030 are the most different intercity and rotation. Tc_000010 and Tc_000012 test suites contain the same 24 classes. OutexTc_000010 has only illuminates "Inca" and OutexTc_000012 have three different illuminates ("Horizon", "Inca" and "TL84").

In this two dataset for each one of different illuminates have nine different rotation angles ($0^\circ, 5^\circ, 10^\circ, 15^\circ, 30^\circ, 45^\circ, 60^\circ, 75^\circ, 90^\circ$) and 20 textures in each one of rotation angles. Texture images of this Outex sets are .RAS format in gray-level.

Tc_000030 is one of the most well known suits of the Outex dataset. It is included 68 texture images in resolution 100dpi and size of $128 \times 128$. Colored image is one of the properties of the Tc-000013. It is suitable for texture analysis in color or gray-scale manners. All Tc-000013 samples are in Inca illumination. Some samples of the Outex Tc-000013 images are shown in Fig. 2.

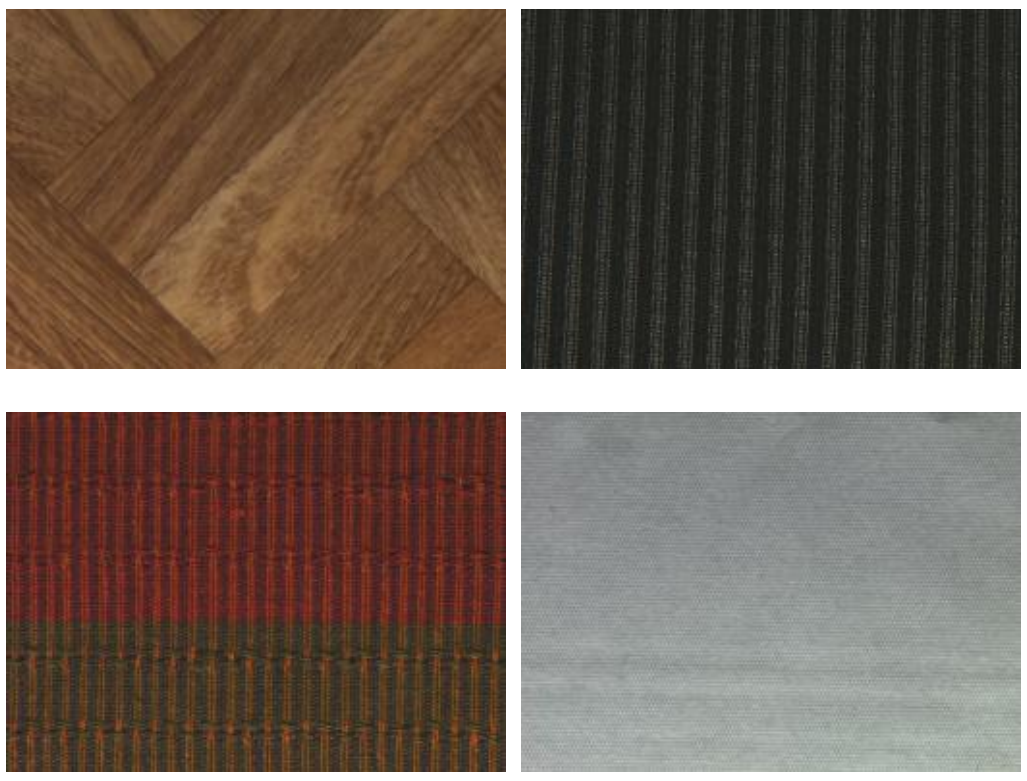

Figure 2. Some examples of the Outex Tc-000030 dataset

*C. VisTex*

Vision texture dataset (VisTex) was product firstly by MIT University [9]. VisTex database contains color texture images. Texture images that are representative of real world conditions, is the main goal of VisTex. Homogeneous texture images can be found in VisTex, but it also contains some real word scenes of multiple textures. All of the images are in 128×128 size. VisTex dataset has 40 classes. Each class has just a one sample. So, in texture classification problems, each texture image should be divided to some non-overlap windows with a same size to provide enough samples for each class. Image windowing may be a good idea here because most of the VisTex image senses are real, so the cropped windows are not same absolutely. Some samples of the VisTex dataset is shown in Fig. 3.

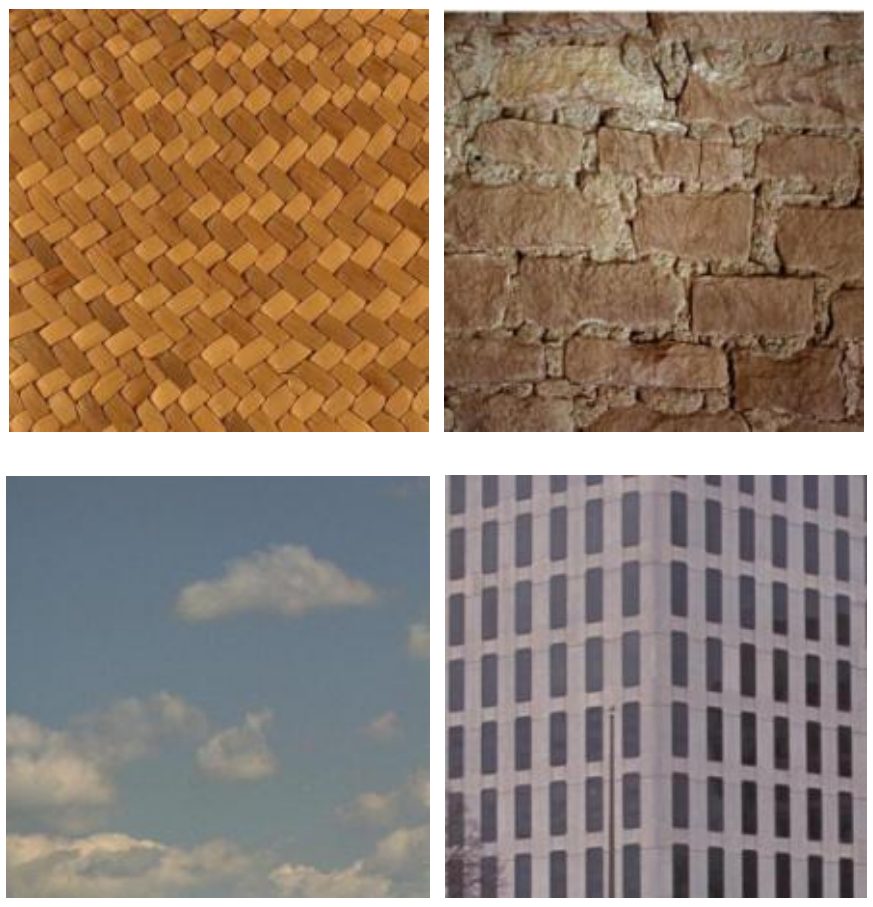

Figure 3. Some Examples of the VisTex Dataset

### D. *Trunk12*

Most of the objects or products have absolutely different textures. So, texture classification in different objects or products may not evaluate the analysis power of texture descriptors accurately. In many cases such as tree, cell, flower, fruit, textile, stone and so on, same objects may have different classes with different textures. In this cases texture images are not very different. Hence, texture classification is not a simple problem. Trunk12 [10] is a texture datasets of 12 different tree barks in different species. Trunk12 consist of 393 samples in 12 classes. The class samples are not balanced. Each class has some samples between 30 to 45 images. All of the images are in 3000 × 4000 size with high resolution. Local rotation and same zoom are other properties of Trunk12. Some examples of Trunk12 dataset is shown in Fig. 4.

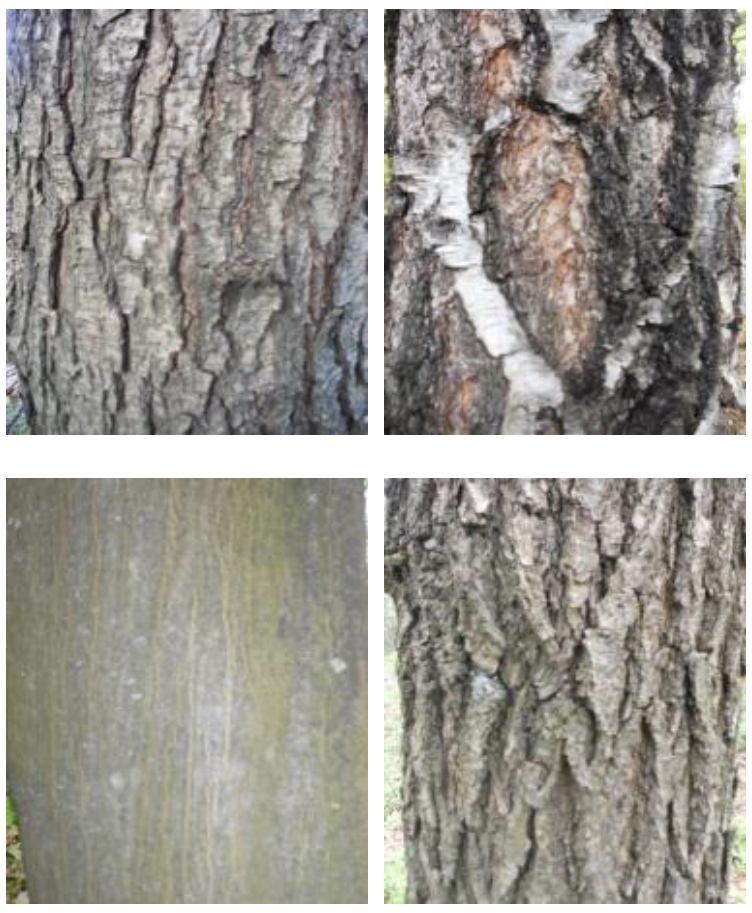

Figure 4. Some Examples of the Trunk12 Dataset

There are some other texture image datasets such as UIUC, FAD, BarkTex, KTH-TIPS, etc. As discussed above, near all of the texture image datasets can be categorized in two groups titled general-based and product-based.

In the general-based datasets, such as Brodatz, Outex and VisTex, texture classes are different in terms of content, object or product. In the product-based datasets, all of the samples are from a unique product, but in different classes such as Trunk12, BarkTex, etc. A good review on texture image analysis methods and texture image datasets is proposed in [11].

## III. PROPOSED BENCHMARK DATASET

As it is discussed above, each abnormality like hole, crash, color variation, is considered as defect in inspection systems. Visual inspection is a main phase in producing process of many kinds of products such as leather, stone, textile, metal, wood, etc. Usually surface abnormalities disturb the primary texture of the surface. So, there is high relation between texture and defect detection problems. For example, in Fig. 5-a, defect-less image of fabric textile is shown. In Fig. 5-b, an example of defected fabric textile is shown. As it is shown, in defect parts of the Fig. 5-b, the primary texture is disturbed.

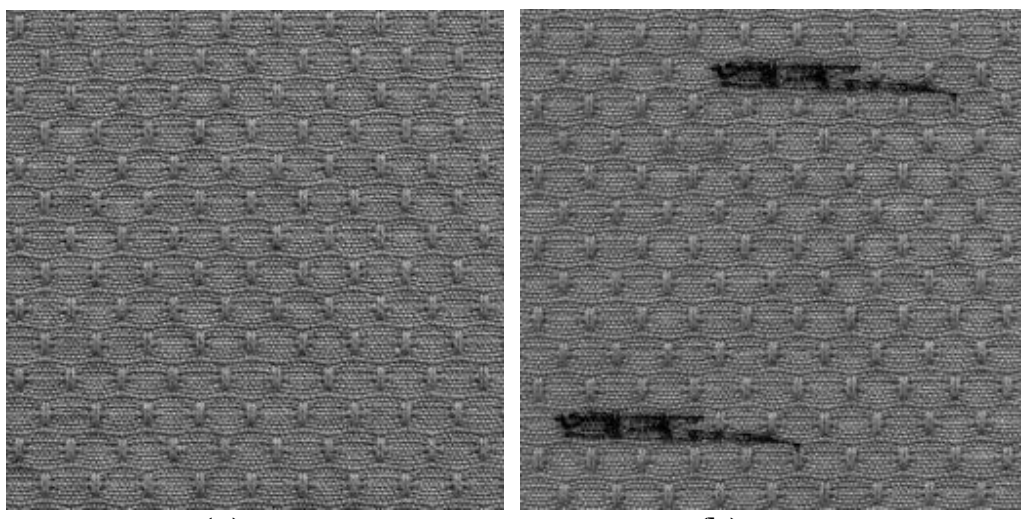

(a) (b)

Figure 5. Some examples of defect-less and defected fabric textile

In this reasons, a dual purpose benchmark dataset is proposed in this paper for texture image analysis and surface defect detection. The title of the proposed dataset is Stone Texture Dataset (STI Dataset). It can be downloaded in [12]. The main product of the collected STI benchmark dataset is stone because of the following reasons:

- In some kinds of stones, repetitive textures are little in size and some other is bigger. This feature makes texture analysis more challengeable. So, texture descriptors should be multi resolution to evaluate good performance on STI dataset.
- On other hand, the visual scenes of some kinds of stones are very same. In another cases, the visual variation of primary texture is more.
- Stone inspection is an important phase of stone quality process. So, visual defect detection on stones is a very popular research topic in image processing and computer vision.

According to these reasons, 60 image were captured of 4 different kinds stone titled hatchet, marble, orange travertine and creamy travertine. All of the images were captured with a digital camera with resolution of 0.2 mm/pixel. The proposed STI benchmark dataset have some properties as follows, which make it more near to real world.

- The size of images is not same.
- For all stone kinds, defect-less and defected samples are included.
- The zoom distance in images is not same. So, in a same kinds, the same textures have different size.
- All of the images are in JPEG format with resolution 72 dpi.
- The local rotation and different poses are other properties of the proposed dataset.
- Images are collected under different illuminations and angles.

### *A. The Structure of the Proposed Dataset for Texture Analysis*

In the complete STI dataset, 20 samples are fully
defect-less which is known as reference samples. five defect-less samples for 4 different kinds of stones. In this respect, a sub set of dataset can be considered as texture image dataset with 5 samples of 4 texture classes. The samples of the proposed texture image dataset are named in the Table 1. An example of each 4 different texture classes of STI is shown in Fig. 6.

Table 1. The structure of STI texture analysis dataset samples

| **Class** | **Hatchet** | **Creamy Travertine** | **Marble** | **Orange Travertine** |
|---|---|---|---|---|
| **Sample Names** | HR1 | CTR1 | MR1 | OTR1 |
| | HR2 | CTR2 | MR2 | OTR2 |
| | HR3 | CTR3 | MR3 | OTR3 |
| | HR4 | CTR4 | MR4 | OTR4 |
| | HR5 | CTR5 | MR5 | OTR5 |
| **Total** | 5 Samples | 5 Samples | 5 Samples | 5 Samples |

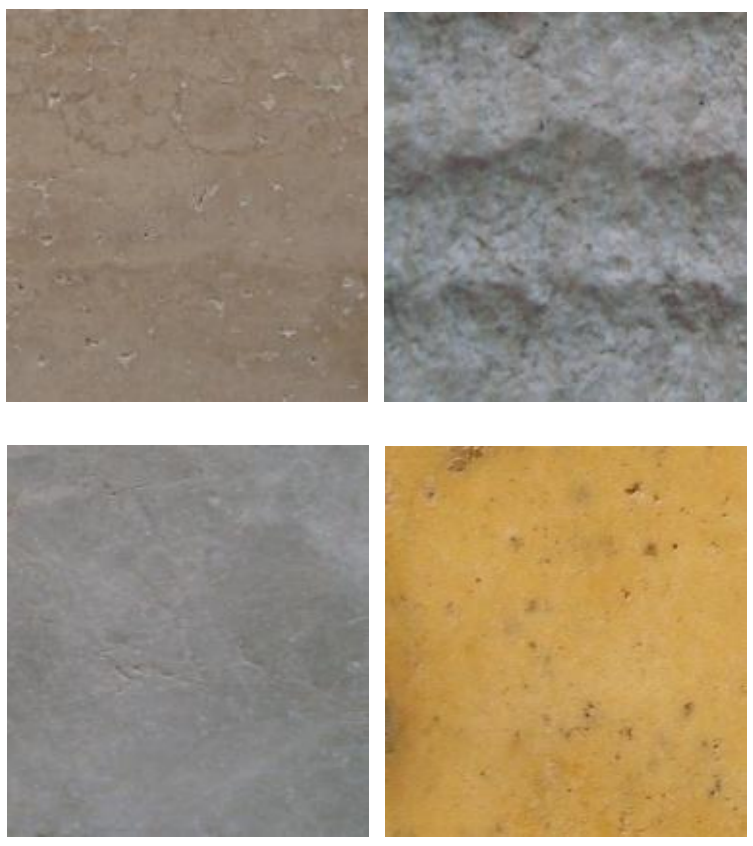

Figure 6. Examples of 4 texture classes of STI dataset

### *B. The Structure of the Proposed Dataset for Surface Defect Detection*

In visual inspection systems, usually defect-less and defected samples should be considered jointly to evaluate the performance. In this respect, 60 images are captured for 4 classes. Foe each class, 5 images are defect-less and 10 images are defected. The samples of the proposed surface defect detection dataset are shown in Table 2 with the real names.

Table 2. The structure of STI defect detection dataset samples

| Class | Hatchet | Creamy Travertine | Marble | Orange Travertine |
|---|---|---|---|---|
| **Train Set Sample Names (Defect-less)** | HR1 | CTR1 | MR1 | OTR1 |
| | HR2 | CTR2 | MR2 | OTR2 |
| | HR3 | CTR3 | MR3 | OTR3 |
| | HR4 | CTR4 | MR4 | OTR4 |
| | HR5 | CTR5 | MR5 | OTR5 |
| **Test Set Sample Names (Defected)** | HD0 | CTD0 | MD0 | OTD0 |
| | HD1 | CTD1 | MD1 | OTD1 |
| | HD2 | CTD2 | MD2 | OTD2 |
| | HD3 | CTD3 | MD3 | OTD3 |
| | HD4 | CTD4 | MD4 | OTD4 |
| | HD5 | CTD5 | MD5 | OTD5 |
| | HD6 | CTD6 | MD6 | OTD6 |
| | HD7 | CTD7 | MD7 | OTD7 |
| | HD8 | CTD8 | MD8 | OTD8 |
| | HD9 | CTD9 | MD9 | OTD9 |
| **Total** | 15 Samples | 15 Samples | 15 Samples | 15 Samples |

Some samples of the proposed STI dataset are shown in the Fig. 7. To use the STI dataset, please cite the related paper in reference [13].

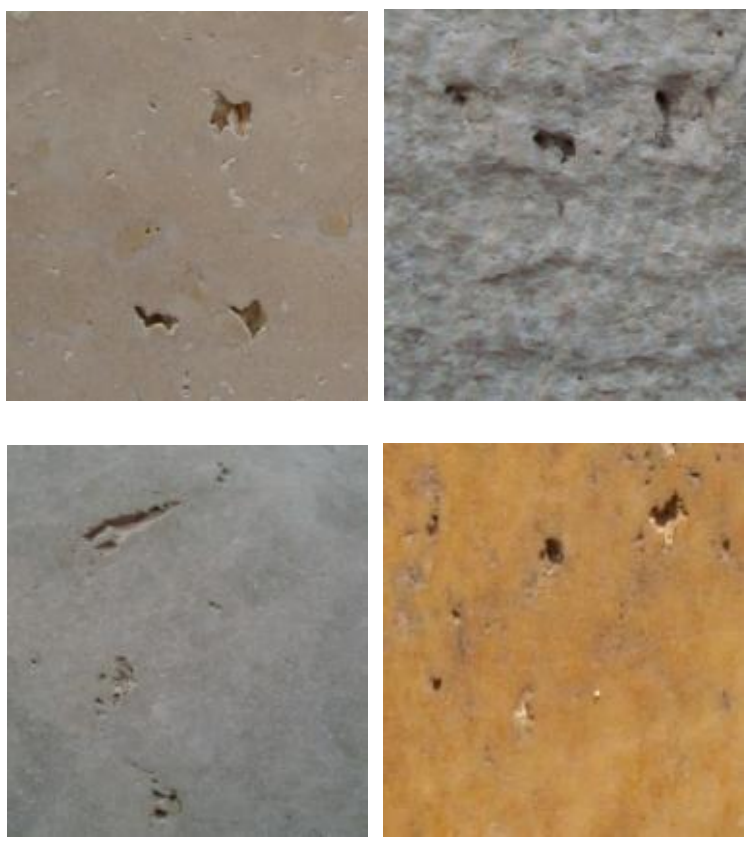

Figure 7. Examples of defected images of STI dataset

## IV. RESUTLS

First time, a demo of the proposed STI dataset with 3 classes, was used in [13], to evaluate the performance of porosity detection system. The results in [13], show that STI dataset is most near to the real world samples.
A more completed version if STI dataset was used in [14], to evaluate the performance of a surface defect detection system. The results of applying some efficient state-of-the-art texture descriptors on STI dataset are shown in table 3. As it shown in table 3, STI dataset is more challengeable for texture analysis and defect detection approaches than many other well known datasets. Some samples of applying defect detection approaches on STI dataset samples are shown in Fig. 8.

Table. 3. Experimental results of applying some texture descriptors on a subset of STI dataset

| Class / Texture Descriptor | Creamy Travertine | Hatchet | Orange Travertine | Average |
|---|---|---|---|---|
| **1DLBP [6]** | 95.60 | 96.2 | 95.7 | 95.83 |
| **MLBP16,2 [15]** | 93.67 | 92.27 | 94.43 | 93.45 |
| **1DLBP + SSR [13]** | 97.33 | 98.06 | 95.82 | 97.07 |
| **NrCLBP [14]** | 96.20 | 97.31 | 95.92 | 96.47 |
| **MBP [16]** | 92.36 | 90.17 | 91.25 | 91.26 |

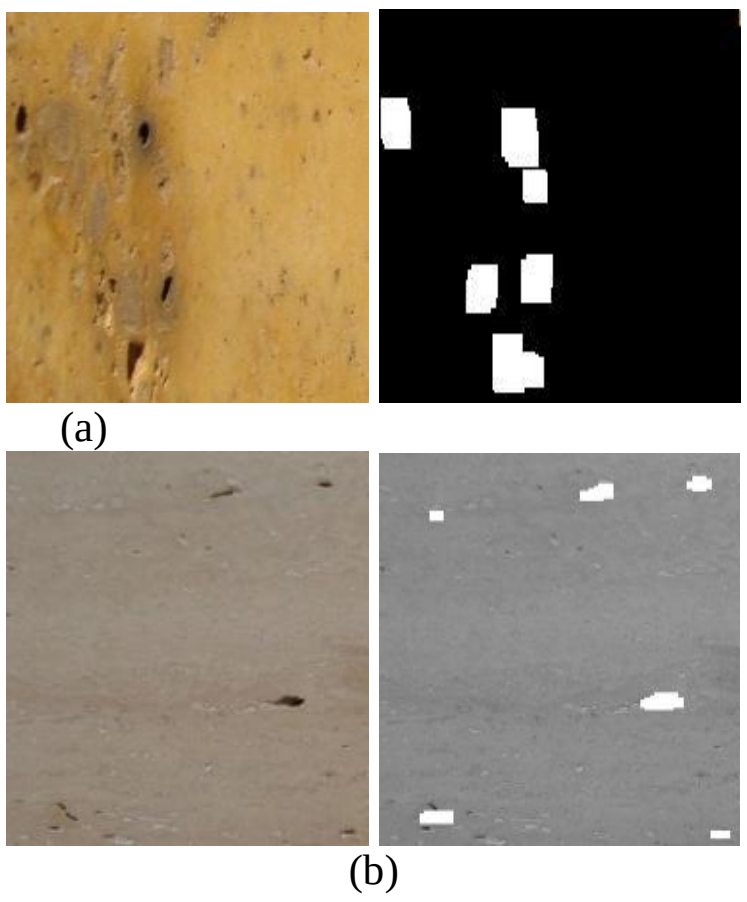

Figure 8. Experimental results of applying texture descriptors on STI texture images for surface defect detection
(a) Noise Resistant Local Binary Patterns (NrCLBP)
(b) One Dimensional Local Binary Patterns (1DLBP)

## V. CONCLUSION

The main aim of this paper was to propose a dual purpose dataset for texture image analysis and surface defect detection jointly. In this respect, STI dataset was proposed with main focus on stone texture images. STI dataset consist of 60 different images of four kinds of stones. A sub-set of STI with 20 samples is designed for texture image analysis and the full set of 60 images is developed as benchmark dataset for surface defect detection. The proposed STI dataset have some unique properties which make it more usable and efficient than many other state-of-the-art datasets.

- Dual purpose is the main advantage of the STI dataset
- Local rotation in captured images
- Different zoom rates in captured images
- Local and global repetitive textures, because of natural features of stone kinds.
- Low-size and high-size repetitive textures jointly
- Different textures with same human visual scene
- Low-size and high-size defects
- Different kinds of abnormalities in color, texture and shape.

To use the STI dataset, please cite the related paper in reference [13].

## Authors' Profiles

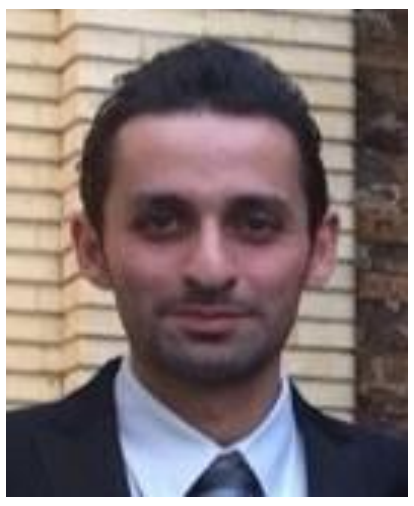

**Shervan Fekri-Ershad** received his B.Sc. degree in computer hardware engineering from Islamic Azad University of Najaf-Abad, Iran in 2009. He received his M.Sc. & Ph.D. degrees from International Shiraz University, Iran in 2012–2016, majored in Artificial Intelligence. He joined the faculty of computer engineering at najafabad branch, Islamic azad university, najafabad, Iran as a staff member (assistant professor) in 2012. His research interests are image processing applications includes visual inspection systems, visual classification, texture analysis, surface defect detection and etc.